\documentclass{Interspeech}

\usepackage{amsmath,amssymb}
\usepackage[table]{xcolor}
\usepackage{multirow}
\RequirePackage{booktabs}
\usepackage{bbm}
\usepackage{cite}
\usepackage{breqn}

\interspeechcameraready

\title{Efficient Trie-based Biasing using K-step Prediction for Rare Word Recognition}

\author[]{Chin Yuen}{Kwok}
\author[]{Jia Qi}{Yip}

\affiliation[nocounter]{College of Computing and Data Science}{Nanyang Technological University}{Singapore}
\email{kwok0062@e.ntu.edu.sg}
\keywords{speech recognition, human-computer interaction, computational paralinguistics}

\usepackage{comment}

\begin{document}

\maketitle

Contextual biasing improves rare word recognition of ASR models by prioritizing the output of rare words during decoding. A common approach is Trie-based biasing, which gives ``bonus scores" to partial hypothesis (e.g. ``Bon") that may lead to the generation of the rare word (e.g. ``Bonham"). If the full word (``Bonham") isn’t ultimately recognized, the system revokes those earlier bonuses. This revocation is limited to beam search and is computationally expensive, particularly for models with large decoders. To overcome these limitations, we propose adapting ASR models to look ahead and predict multiple steps at once. This avoids the revocation step entirely by better estimating whether a partial hypothesis will lead to the generation of the full rare word. By fine-tuning Whisper with only 10 hours of synthetic data, our method reduces the word error rate on the NSC Part 2 test set from 30.86\% to 12.19\%.

\begin{abstract}
     
\end{abstract}

\section{Introduction}

Accurate recognition of rare words is essential for ASR systems, as real-world speech often includes new, domain-specific, or underrepresented terms. However, the rare occurrence of these words in training data limits ASR performance. To address this, synthetic training data generated via text-to-speech (TTS) systems can be used as a substitute \cite{Zheng2020UsingSA}. By generating audio samples from sentences containing the target rare words, TTS enables the creation of training datasets tailored to specific contexts. This approach alleviates the need for extensive real-world recordings and allows the ASR model to automatically learn rare words by further adapting to the synthetic data.

\begin{figure}[t]
  \centering
  \includegraphics[width=0.8\linewidth]{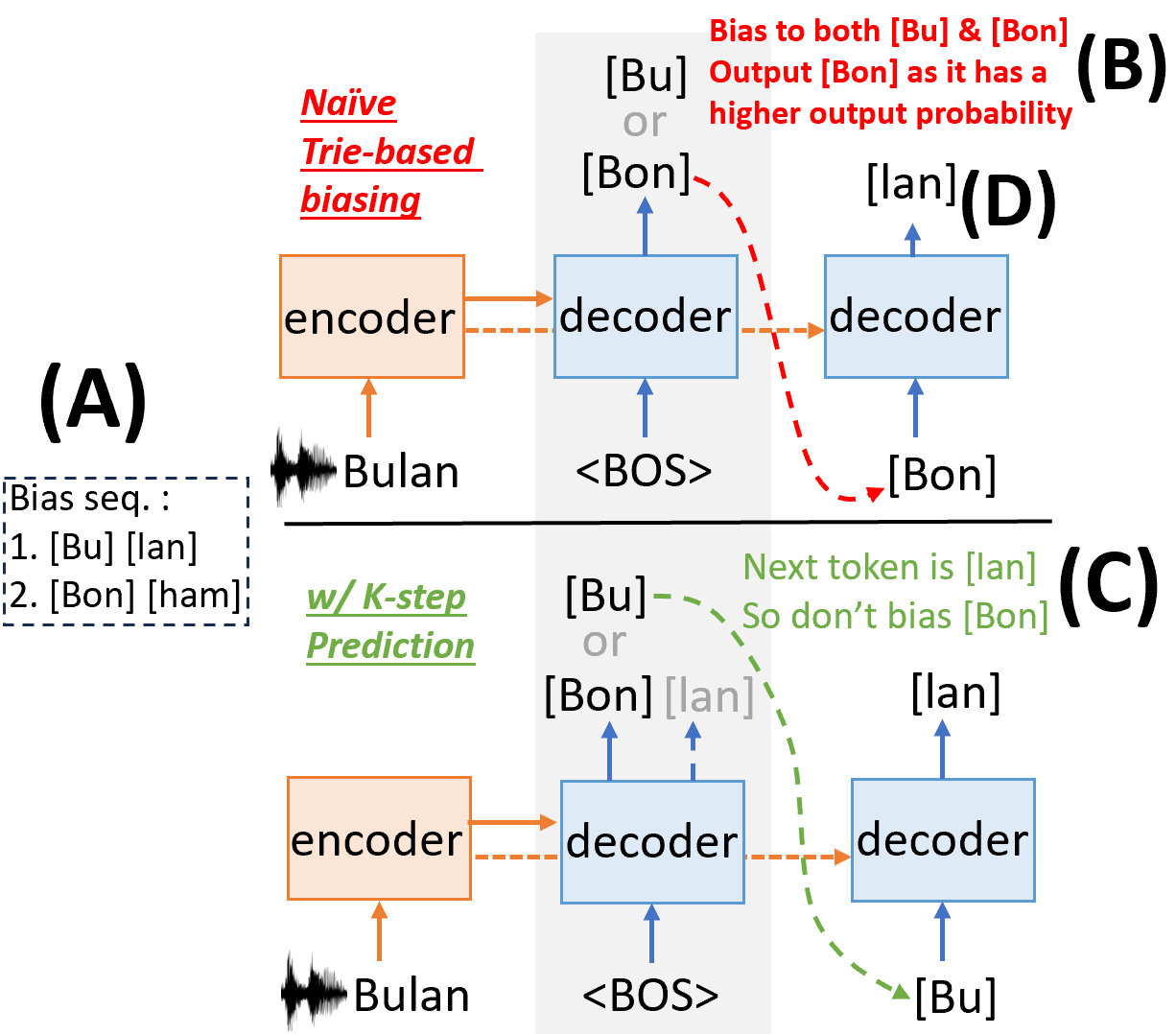}
  \caption{Overview of incorporating $K$-step prediction for Trie-based biasing using Whisper. Given that the input audio says ``Bulan" and A) the biased words are ``Bulan" and ``Bonham", B) naive Trie-based biasing will bias both tokens ``[Bon]" and ``[Bu]" with equal rewards. If Whisper wrongly output a higher probability for ``[Bon]", the output token will be ``[Bon]". C) After adapting Whisper to perform $K$-step predictions on the rare words, the model can predict whether the future token is ``[ham]" ($K$-step prediction is cheap as it does not require a full decoder pass to predict the future token). As the future token is ``[lan]" instead of ``[ham]", the token ``[Bon]" will not be biased. This reduces the chance of biasing to the wrong rare word.
}
  \label{fig:mtp_trie}
\end{figure}

To further improve the accuracy of rare word recognition after an ASR model is adapted to the synthetic data, contextual biasing \cite{nigmatulina2023implementing,xu2023adaptive,harding23_interspeech,sathyendra2022contextual,naowarat23_interspeech,shamsian2024keyword,liu2020contextualizing,sun23e_interspeech,futami2024phoneme} is used to prioritize the output of rare words during decoding. One approach is pronunciation-based post-processing \cite{le2020g2g,le2021deep}, which corrects misrecognized rare words by leveraging phonetic similarities. In this approach, the FT model decodes the real train set to identify substitutions of rare words with phonetically similar ones, creating a mapping. During inference, these substitutions are reversed to restore rare words. Alternatively, methods such as contextual adapters \cite{gong2024contextual} and TCPGen \cite{sun2023contextualbiasingremaineffective} focus on decoder-side adaptation \cite{kwok2024continual,kwok2024continual_2}, which integrate a biasing component directly into end-to-end ASR system to improve recognition of contextually relevant words.

A more classical inference-based approach is Trie-based biasing \cite{Zhao2019ShallowFusionEC}, which assigns higher scores to partial hypotheses (e.g. ``Bon") that may lead to the generation of the rare word (e.g. ``Bonham").  This approach is particularly useful in scenarios where specific word lists, such as names in a contact directory or locations in navigation systems, can be leveraged. However, since the rare words in synthetic audio may have slightly different pronunciations than the real audio, the ASR model adapted to the synthetic audio may not be able to distinguish the beginning segments of the rare words (e.g. ``Bon" in ``Bonhan" and ``Bu" in ``Bulan") if they have similar pronunciations as shown in Fig. \ref{fig:mtp_trie}B. This will cause errors when a partial hypothesis, e.g. ``[Bon]", is rewarded by Trie-based biasing but ultimately fails to produce the full subword unit sequence``[Bon] [ham]" as shown in Fig. \ref{fig:mtp_trie}D.

Traditionally, this issue is addressed through the reward revocation step \cite{Pundak2018DeepCE} in Trie-based biasing. During decoding, Trie-based biasing will assign higher scores to partial hypotheses that may lead to the generation of the rare word. However, if the full rare word is not ultimately generated, these rewards are revoked. This revocation is restricted to beam search and incurs high computational costs, particularly for models with large decoders, reducing the efficiency of Trie-based biasing in real-time ASR applications.

\begin{figure}[t]
  \centering
  \includegraphics[width=0.8\linewidth]{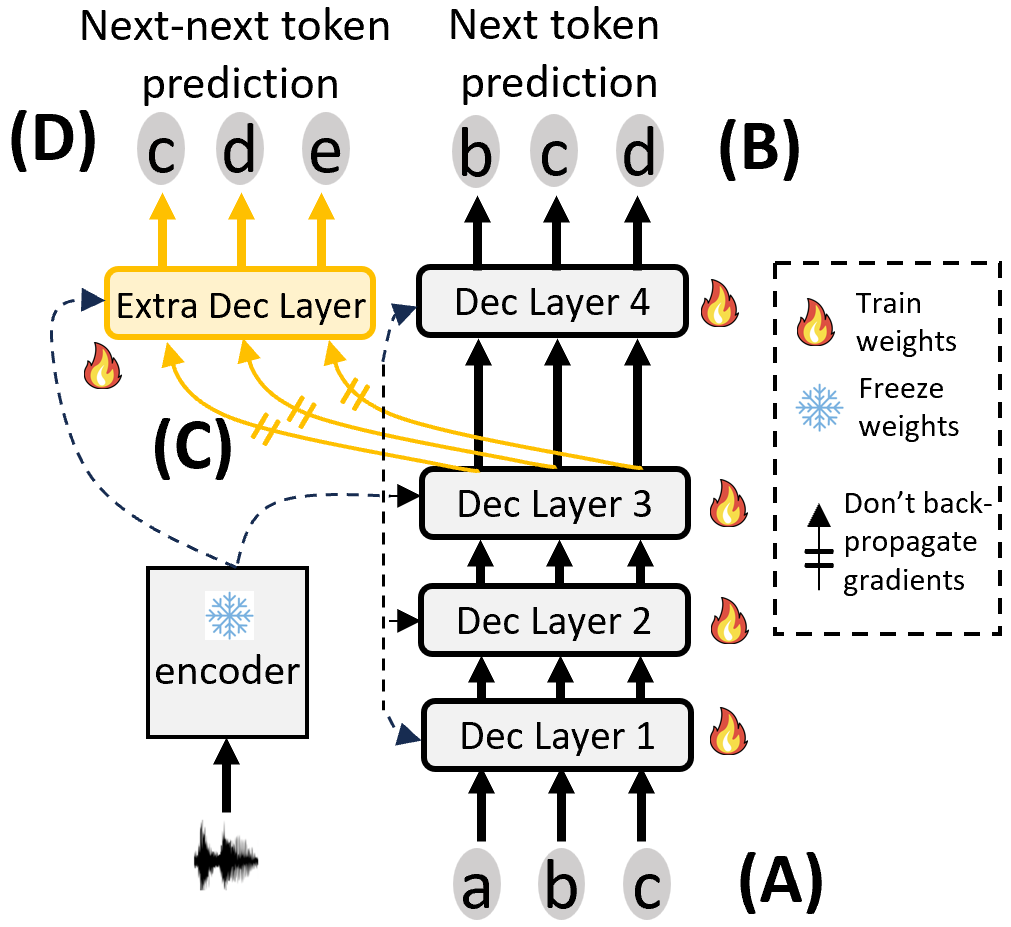}
  \caption{Overview of our Whisper architecture with $K$-step prediction. Assume $K=2$ and only one extra future token (next-next token) is predicted, and assume the decoder has four layers. Given a sequence ``a b c d e", A) the decoder is trained to take in ``a b c" to B) generate ``b c d" in an auto-regressive manner for next-token prediction. To extend Whisper with two-step prediction capabilities, C) an extra decoder layer is added right before the last decoder layer to output D) next-next token prediction ``c d e". Only the encoder weights are frozen during training. The gradients from the extra decoder layer will not back-propagate to the original decoder to ensure that the fine-tuning process of the original decoder is not interfered.
}
  \label{fig:mtp_architecture}
\end{figure}

To mitigate these challenges, we extend Trie-based biasing \cite{Zhao2019ShallowFusionEC} and propose a $K$-step prediction approach to better recognize rare words without the need for reward revocation, inspired by the historical ASR look-ahead and acoustic fast match methods \cite{ortmanns2000look,liu2001new}. Specifically, we adapt ASR models using synthetic training data to enable the models to predict multiple future tokens at once as shown in Fig. \ref{fig:mtp_architecture}. By conditioning the models to anticipate whether future tokens will form a complete rare word, we enhance the ability of Trie-based biasing to assign appropriate score rewards to hypotheses as shown in Fig. \ref{fig:mtp_trie}C. Our method serves as an alternative to reward revocation and is compatible with greedy search, eliminating the need for costly beam search while enhancing recognition accuracy.

Our contributions are threefold. We show that: 1) we can expand Whisper \cite{Radford2022RobustSR} to effectively perform $K$-step prediction on rare words by only adapting to 10 hours of synthetic audio data, which are generated from sentences containing the rare words. 2) $K$-step prediction allows ASR models to look ahead and predict multiple steps at once. This avoids the revocation step of Trie-based biasing entirely by better estimating whether a partial hypothesis will lead to the generation of the full rare word. 3) Our Trie-based biasing with $K$-step prediction is more effective than naive Trie-based biasing when more distractors (irrelevant words) are included in the list of words to bias.

\section{Method}
\label{sec:method}

\subsection{Beam Search}

Given acoustic observations $x$ from an audio sample and its corresponding transcript represented as a sequence of subword units $y_{1:L} = (y_1,\dots,y_L)$, end-to-end ASR models are trained to generate the subword units in an autoregressive manner. Specifically, these models estimate the posterior probability of each subword unit conditioned on the previously predicted units and the acoustic input, formulated as $P(y_n|y_{1:n-1},x)$. The goal is to find the optimal $y$ given $x$:
\newcommand\mydots{\hbox to 1em{.\hss.\hss.}}
\begin{equation}
y^{*} = \underset{y}{\arg\max} P(y_1|,x)P(y_2|y_1,x)\dots{}P(y_n|y_{1:n-1},x)
\label{eq:ar}
\end{equation}

As each of the posterior probabilities in Eq. \ref{eq:ar} depends on the previous decoded units, it requires brute-force to compute the posterior probabilities through all possible sequences of $y$ to find $y^*$. To reduce computation cost, beam search is used to approximate Eq. \ref{eq:ar} while limiting the search space. 

Beam search \cite{kwok2024improved} proceeds as follows. Let $J$ denote the beam size and  $V$ the vocabulary size of the ASR model. The process begins with a single sequence containing only the beginning-of-sentence (BOS) token. At each decoding step, the search expands the current sequences by appending the top $J$ tokens with the highest probabilities from the full set of $V$ possible tokens, generating $J$ candidate sequences. This expansion continues iteratively, where each sequence is again extended by selecting the top $J$ tokens, resulting in $J^2$ candidate sequences. To maintain computational efficiency, only the top $J$ sequences with the highest probabilities are retained after each step. Decoding of a sequence concludes when it is extended with an end-of-sentence (EOS) token, and the beam search process continues until all sequences in the beam terminate with EOS.

\subsection{Contextual Biasing}

Let $C$ be the event that $x$ contains words that belong to a list of rarewords. For simplicity, we omit the auto-regressive notations in Eq. \ref{eq:ar} and only focus on $P(y|x)$. The posterior probability is decomposed as:
\begin{dmath}
\label{eq:decomp1}
    P(y|x) = P(y|C,x) P(C|x) + P(y|\overline{C},x) P(\overline{C}|x)
\end{dmath}

To perform context biasing, a bias list is generated for each audio sample in the test set indicating what words are likely to appear in the audio. Then, a new posterior probability $Q(y|x)$ will be estimated for the test set distribution:
\begin{dmath}
\label{eq:decomp2}
    Q(y|x) = Q(y|C,x) Q(C|x) + Q(y|\overline{C},x) Q(\overline{C}|x)
\end{dmath}

Compared to the train set distribution, assume that $Q(y|C,x)=P(y|C,x)$ and $Q(y|\overline{C},x)=P(y|\overline{C},x)$, and assume the only difference between the distributions of the test and train set is that $Q(C|x)>P(C|x)$. Then $Q(y|x)$ is:
\begin{dmath}
\label{eq:decomp3}
    Q(y|x) = P(y|C,x) Q(C|x) + P(y|\overline{C},x) Q(\overline{C}|x)
\end{dmath}

Compared to $P(y|x)$, $Q(y|x)$ in Eq. \ref{eq:decomp3} have a larger multiplier on $P(y|C,x)$ and a smaller multiplier on $P(y|\overline{C},x)$ as $Q(C|x)>P(C|x)$ and $Q(\overline{C}|x)<P(\overline{C}|x)$. This is because the bias list is generated for the test set, so the event $C$ is more probable for $x$ sampled from the test set distribution. This means that $P(y|C,x)$ will have more influence on $Q(y|x)$ for context biasing purpose, and less influence on $P(y|x)$.

\subsection{Trie-based Biasing}

Using the auto-regressive notations, the new posterior probability in Eq. \ref{eq:decomp3} is:
\begin{dmath}
Q(y_n|y_{1:n-1},x) = \\P(y_n|C,y_{1:n-1},x)Q(C|x) + P(y_n|\overline{C},y_{1:n-1},x)Q(\overline{C}|x)
\end{dmath}

\noindent in which Trie-based biasing \cite{Zhao2019ShallowFusionEC} provides a simple estimation of the four terms on the right hand side. The four estimations are:
\begin{equation}
    P(y_n|C,y_{1:n-1}x) \approx P(y_n|y_{1:n-1}x) \times \mathbbm{1}(y_{1:n})
\end{equation}
\begin{equation}
    P(y_n|\overline{C},y_{1:n-1}x) \approx P(y_n|y_{1:n-1}x) \times (1-\mathbbm{1}(y_{1:n}))
\end{equation}
\begin{equation}
    Q(C|x) \approx \alpha{} P(C|x)
\end{equation}
\begin{equation}
    Q(\overline{C}|x) \approx \beta{} P(\overline{C}|x)
\end{equation}

\noindent where $\alpha$ and $\beta$ are scalars such that $\alpha{} P(C|x) + \beta{} P(\overline{C}|x) = 1$. $\mathbbm{1}(y_{1:n})$ is an indicator function that outputs one if and only if $y_n$ leads to the full or partial generation of a biased word. Otherwise, $\mathbbm{1}(y_{1:n})$ outputs zero.

Given the four estimations, the new log-probability score is then:
\begin{dmath}
\label{eq:trie_score}
\log Q(y_n|y_{1:n-1},x)=\\
    \begin{cases}
        \log P(y_n|y_{1:n-1},x) + \log \alpha & \text{if}\,\,\, \mathbbm{1}(y_{1:n}) = 1\\
        \log P(y_n|y_{1:n-1},x) + \log \beta & \text{otherwise}
    \end{cases}
\end{dmath}

\noindent where $\log \alpha$ is a positive real number and $\log \beta$ is a negative real number. In other words, to compute the new log-probability score, a positive reward is added to the original log-probability score if $y_n$ extends a rareword sequence, and a negative penalty is added otherwise. In practice, the reward and penalty is substracted by $\log \beta$ to become $\lambda = \log \alpha - \log \beta$ and zero respectively for convenience. This will not change the outcome of the decoding as the ranking of the scores is not changed after the subtraction.

\subsection{Reward Revocation and $K$-step Prediction}

As the estimation by Trie-based decoding may not be accurate, the reward must be revoked if overbiasing occurs. As shown in Fig. \ref{fig:mtp_trie}, Given that the input audio says “Bulan” and the biased words are ``Bulan” and ``Bonham", naive Trie-based biasing will bias both tokens ``[Bon]" and ``[Bu]" with equal rewards. If the ASR model wrongly output a higher probability for ``[Bon]", the output token will be ``[Bon]". However, as the output token following “[Bon]” is ``[lan]" instead of ``[ham]", it means that the biasing of the rare word failed and the reward must be revoked to discourage the model to retain this candidate sequence during beam search.

However, beam search is computationally expensive for models with large decoders. To address this, as shown in Fig. \ref{fig:mtp_trie}C, we propose adapting the ASR model to perform $K$-step prediction, e.g. two-step prediction, on the rare words such that the model can predict whether the token in one future step is “[ham]” without explicitly decoding for one more step. Then we will set the score reward for ``[Bon]" to zero if the future token is not “[ham]”. Specifically, we replace the term $\mathbbm{1}(y_{1:n})$ in Eq. \ref{eq:trie_score} as $\mathbbm{1}^{\prime{}}(y_{1:n})$, which is an indicator function that outputs one if and only if $y_n$ leads to the full or partial generation of a biased word and the tokens with the top $\mu$ highest probabilities in the K-step prediction can continue the generation of the word. Otherwise, $\mathbbm{1}^{\prime{}}(y_{1:n})$ outputs zero.

This approach enables a more effective assignment of scores to partial hypotheses by predicting whether the $K-1$ future tokens can continue to extend the subword unit sequence for the rare word and eliminates the need for reward revocation.

\section{Experiment Setup}
\label{sec:exp}

\subsection{Dataset and metrics}
\label{ssec:dataset}

Experiments were conducted on the National Speech Corpus Part 2 (NSC-Part-2) \cite{koh2019building}, a Singapore English dataset comprising 13K unique utterances of people asking for directions, including road names and addresses. A development set of 500 utterances was held out. This dataset was selected because synthetic audio has been publicly released \cite{yuen2023asr}, with approximately 15\% of words being rare, compared to only 2\% in LibriSpeech \cite{zen2019libritts}, making it more suitable for rare word evaluation.

We used 20 hours of real training audio, 10 hours of synthetic audio (VITS-SPKSET1 \cite{yuen2023asr}), and a 2-hour real test set. The synthetic audio was generated using the VITS text-to-speech model \cite{kim2021conditional} from coqui-ai\footnote{\url{https://github.com/coqui-ai/TTS}}, trained on the VCTK corpus \cite{Yamagishi2019CSTRVC}, using the same transcripts as the real training set. Despite accent differences between VCTK and NSC-Part-2, the synthetic audio was effective for text domain adaptation.

Similar to Sun et al. \cite{sun2023contextualbiasingremaineffective}, we treat words that do not exist in the 10K common word list \footnote{https://github.com/first20hours/google-10000-english/blob/master/google-10000-english.txt} as rare words. In total, there are 3856 and 1106 rare words in the train and test sets respectively. In this setup all the testset rarewords are considered OOV words as we never fine-tune on any real audio data containing the words. During inference, a list of biasing words for each utterance was obtained by collecting rarewords that appear in the utterance and adding $N$ distractors, which are randomly sampled from the set of all possible rarewords from the train set.

To evaluate our method, two additional metrics were used to evaluate system performance besides word error rate (WER): the biased word error rate (B-WER) and the unbiased word error rate (U-WER) \cite{le2021contextualized}. B-WER is computed as the WER on words that appear in the bias list. In this metric, insertion errors are included only if the inserted word is also in the bias list. Conversely, U-WER measures WER on all other words and errors not associated with the bias list.

\subsection{Implementation details}
\label{ssec:implementation}

We implement our methods based on the popular SpeechBrain \cite{ravanelli2021speechbraingeneralpurposespeechtoolkit} toolkit. We adapt whisper-small and whisper-large-v2 with vanilla fine-tuning (FT) for 2 epochs and set the train batch size to 6. We set the learning rate to $0.005$. We use the AdamW optimizer \cite{loshchilov2017decoupled} with a variant\footnote{\url{https://speechbrain.readthedocs.io/en/latest/\_modules/speechbrain/nnet/schedulers.html\#NewBobScheduler}} of the ReduceLROnPlateau learning rate (LR) scheduler. The encoder weights are frozen to prevent overfitting \cite{kwok2024low,kwok2025extending}. We sweep through the hyper-parameters to optimize WER. Unless otherwise specified, greedy decoding is used for all experiments. We set $K=2$, $\lambda=3$ and $\mu=10$ and perform two-step predictions for trie-based biasing as we empirically find that Whisper has difficulty predicting tokens for $K>2$ steps at once given the limited training data.

\section{Results and Discussions}
\label{sec:results}

\begin{table*}[t]
  \centering
  \caption{\small Contextual biasing for whisper-small. Either the real of synthetic (Syn) version of the NSC-Part-2 train/development set is used for training/hyper-parameter tuning. $N$ is the number of distractors added to the biasing words list. The first block has repeated results as N is irrelevant. KP means $K$-step prediction.}
  \label{tab:exp-librispeech}
  \makebox[\textwidth][c]{
  \begin{tabular}{lcccccccccccc} 
    \toprule
    \multirow{2}{3.5cm}{Method} & \multicolumn{2}{c}{NSC-Part-2} & \multicolumn{3}{c}{N=10} & \multicolumn{3}{c}{N=50} & \multicolumn{3}{c}{N=100} \\ 
    \cmidrule(lr){2-3} \cmidrule(lr){4-6} \cmidrule(lr){7-9} \cmidrule(lr){10-12} 
    & Real & Syn & WER & BWER & UWER & WER & BWER & UWER & WER & BWER & UWER \\
    \midrule
    unadapted & & & 30.86 & 55.80 & 18.03 & 30.86 & 55.80 & 18.03 & 30.86 & 55.80 & 18.03 \\
    Kwok et al. \cite{yuen2023asr} & & \checkmark & 16.5 & - & - & 16.5 & - & - & 16.5 & - & - \\
    FT & \checkmark & & 10.04 & 18.14 & 5.88 & 10.04 & 18.14 & 5.88 & 10.04 & 18.14 & 5.88 \\ 
    FT & & \checkmark & 16.29 & 29.66 & 9.42 & 16.29 & 29.66 & 9.42 & 16.29 & 29.66 & 9.42 \\
    \cmidrule{1-12}
    \multicolumn{1}{l}{\textit{reproduced baselines}}\\
    FT & & \checkmark &\\
    \multicolumn{2}{l}{\quad + Contextual Adapter \cite{gong2024contextual}} & \checkmark & 16.36 & 29.64 & 9.54 & 16.28 & 29.63 & 9.42 & 16.22 & 29.55 & 9.36 \\
    \quad + TCPGen \cite{sun2023contextualbiasingremaineffective} & & \checkmark & 16.29 & 29.66 & 9.42 & 16.29 & 29.66 & 9.42 & 16.29 & 29.66 & 9.42 \\
    \quad + Alter. pron. \cite{le2020g2g} &  & \checkmark & 16.23 & 29.20 & 9.56 & 16.61 & 29.39 & 10.04 & 17.20 & 29.66 & 10.80 \\
    \quad + Alter. pron. \cite{le2021deep} & \checkmark &  & 15.04 & 26.63 & 9.09 & 15.47 & 27.46 & 9.31 & 15.77 & 28.21 & 9.38 \\
    \multicolumn{2}{l}{\quad + Trie-based biasing \cite{Zhao2019ShallowFusionEC}} & \checkmark & 12.56 & 21.12 & 8.16 & 13.61 & 23.03 & 8.77 & 14.57 & 24.91 & 9.25 \\\cmidrule{1-12}
    \multicolumn{1}{l}{\textit{our method}}\\
    FT & & \checkmark & \\
    \quad + Trie-based biasing w/ KP & & \checkmark & \textbf{12.19} & \textbf{20.93} & \textbf{7.71} & \textbf{12.44} & \textbf{21.39} & \textbf{7.84} & \textbf{12.64} & \textbf{21.84} & \textbf{7.91} \\
    \bottomrule
  \end{tabular}
  }
\end{table*}

Table \ref{tab:exp-librispeech} presents the contextual biasing results for whisper-small. Initially, the model performs poorly on NSC-Part-2, with a WER of 30.85\%. Fine-tuning (FT) on real NSC-Part-2 training data substantially improves WER to 10.04\%. As real audio training data may be unavailable, synthetic data can be used instead to enhance rare word recognition. Results show that fine-tuning whisper-small on synthetic data reduces WER to 16.29\%, though a performance gap remains compared to using real audio data. This discrepancy likely stems from synthetic data failing to fully capture real audio distributions, leading to potential overfitting to synthetic artifacts.


To further improve rare word recognition, contextual biasing baseline methods are applied to the synthetic-data fine-tuned model (second block of Table \ref{tab:exp-librispeech}). First, contextual adapters and TCPGen yield no improvement, with TCPGen’s $P^{ptr}(y_i)$ remaining close to zero. This is because the FT model has already overfitted to the NSC-Part-2 synthetic train set and exhibits minimal errors on the train set. Further training of the biasing modules, which are essentially error correction modules, on the same train set will be ineffective.

An alternative baseline is pronunciation-based post-processing, which improves WER from 16.29\% to 15.04\%. However, the method is less effective if synthetic train set is used, likely due to an accent mismatch -- NSC-Part-2 features Singaporean English, while the TTS-generated synthetic data reflects non-Singaporean accents, reducing mapping accuracy.

\begin{figure}[t]
  \centering
  \includegraphics[width=0.6\linewidth]{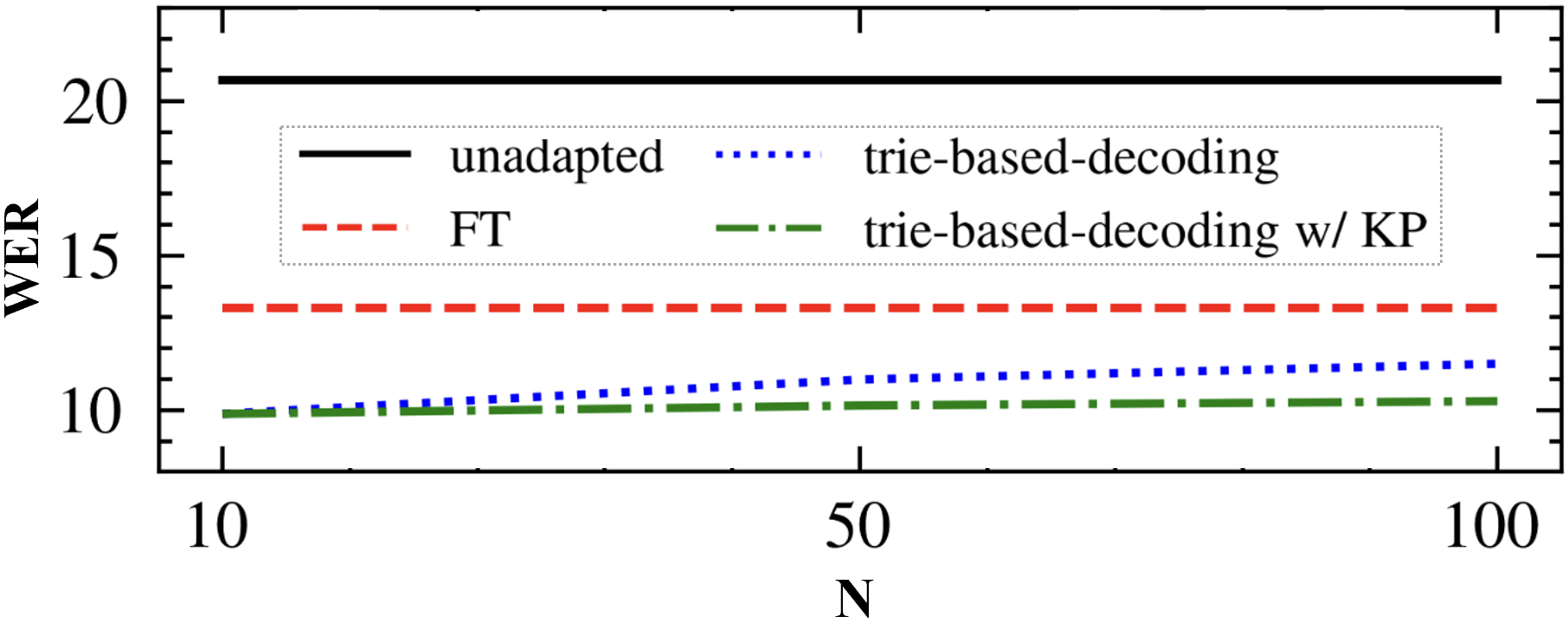}
  \caption{Contextual biasing for whisper-large-v2. Only synthetic data is used for model adaptation. $N$ is the number of distractors added to the biasing words list. KP means K-step prediction.
}
  \label{fig:whisper_large_result}
\end{figure}

Next, the results show that Trie-based biasing achieves the best performance, improving WER from 16.29\% to 12.56\% when the number of distractors $N=10$. However, its effectiveness diminishes as $N$ increases, likely because the method biases to the wrong rare words. Also, score revocation requires computationally expensive beam search and does not work in the greedy decoding scenario. To eliminate the need for reward revocation, we further adapt the FT model for $K$-step prediction capability. As expected, the results in both Table \ref{tab:exp-librispeech} and Fig. \ref{fig:whisper_large_result} show that our method can effectively improve the original Trie-based biasing results, especially when $N$ is large and when the increased number of distractors makes it easier to bias to the wrong rare words.

\begin{figure}[t]
  \centering
  \includegraphics[width=0.8\linewidth]{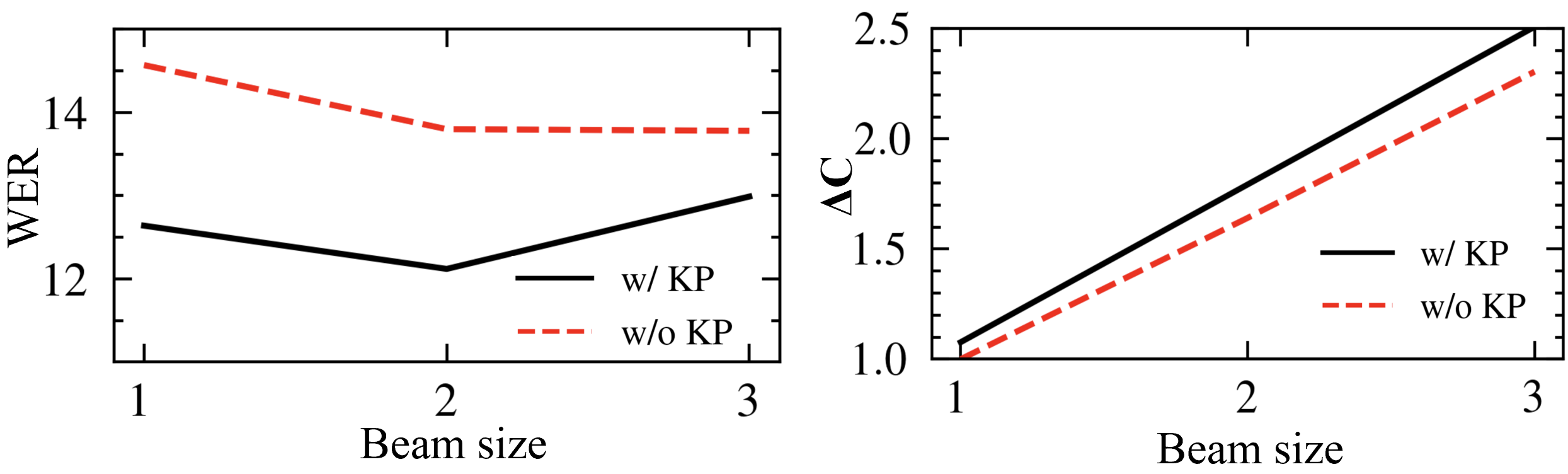}
  \caption{The effect of beam size on the WER performance and computational cost of Trie-based biasing. KP means K-step prediction. $\Delta C$ means the increase in the computational cost relative to decoding with beam size of 1 without KP. $N$ is set to $100$.
}
  \label{fig:cost}
\end{figure}

\subsection{Ablation Study}

We further study the effect of beam size on the WER performance and computational cost of Trie-based biasing as shown in Fig. \ref{fig:cost}. The left plot shows that WER is constantly reduced for beam sizes of 1 and 2, but saturates at beam 3. We hypothesize that this is because the large rewards used for Trie-based biasing are tuned for smaller beam sizes and may cause overbiasing for larger beam sizes. The right plot shows that increasing beam size linearly increases computational cost, and K-step prediction has less influence on the computational cost.

\section{Conclusion}
Our proposed method enhances contextual biasing by enabling ASR models to predict multiple future tokens, eliminating the need for reward revocation in Trie-based biasing. Fine-tuning Whisper with just 10 hours of synthetic data significantly reduced the WER from 30.86\% to 12.19\%, demonstrating the effectiveness of our method in improving rare word recognition.

\section{Acknowledgements}
The computational work for this article was partially performed on resources of the National Supercomputing Centre, Singapore (https://www.nscc.sg).

\bibliographystyle{IEEEtran}
\bibliography{mybib}

\begin{thebibliography}{10}
\providecommand{\url}[1]{#1}
\csname url@samestyle\endcsname
\providecommand{\newblock}{\relax}
\providecommand{\bibinfo}[2]{#2}
\providecommand{\BIBentrySTDinterwordspacing}{\spaceskip=0pt\relax}
\providecommand{\BIBentryALTinterwordstretchfactor}{4}
\providecommand{\BIBentryALTinterwordspacing}{\spaceskip=\fontdimen2\font plus
\BIBentryALTinterwordstretchfactor\fontdimen3\font minus \fontdimen4\font\relax}
\providecommand{\BIBforeignlanguage}[2]{{%
\expandafter\ifx\csname l@#1\endcsname\relax
\typeout{** WARNING: IEEEtran.bst: No hyphenation pattern has been}%
\typeout{** loaded for the language `#1'. Using the pattern for}%
\typeout{** the default language instead.}%
\else
\language=\csname l@#1\endcsname
\fi
#2}}
\providecommand{\BIBdecl}{\relax}
\BIBdecl

\bibitem{Zheng2020UsingSA}
\BIBentryALTinterwordspacing
X.~Zheng, Y.~Liu, D.~Gunceler, and D.~Willett, ``Using synthetic audio to improve the recognition of out-of-vocabulary words in end-to-end asr systems,'' \emph{ICASSP 2021 - 2021 IEEE International Conference on Acoustics, Speech and Signal Processing (ICASSP)}, pp. 5674--5678, 2020. [Online]. Available: \url{https://api.semanticscholar.org/CorpusID:227126398}
\BIBentrySTDinterwordspacing

\bibitem{nigmatulina2023implementing}
I.~Nigmatulina, S.~Madikeri, E.~Villatoro-Tello, P.~Motli{\v{c}}ek, J.~Zuluaga-Gomez, K.~Pandia, and A.~Ganapathiraju, ``Implementing contextual biasing in gpu decoder for online asr,'' \emph{arXiv preprint arXiv:2306.15685}, 2023.

\bibitem{xu2023adaptive}
T.~Xu, Z.~Yang, K.~Huang, P.~Guo, A.~Zhang, B.~Li, C.~Chen, C.~Li, and L.~Xie, ``Adaptive contextual biasing for transducer based streaming speech recognition,'' \emph{arXiv preprint arXiv:2306.00804}, 2023.

\bibitem{harding23_interspeech}
P.~Harding, S.~Tong, and S.~Wiesler, ``{Selective Biasing with Trie-based Contextual Adapters for Personalised Speech Recognition using Neural Transducers},'' in \emph{Proc. INTERSPEECH 2023}, 2023, pp. 256--260.

\bibitem{sathyendra2022contextual}
K.~M. Sathyendra, T.~Muniyappa, F.-J. Chang, J.~Liu, J.~Su, G.~P. Strimel, A.~Mouchtaris, and S.~Kunzmann, ``Contextual adapters for personalized speech recognition in neural transducers,'' in \emph{ICASSP 2022-2022 IEEE International Conference on Acoustics, Speech and Signal Processing (ICASSP)}.\hskip 1em plus 0.5em minus 0.4em\relax IEEE, 2022, pp. 8537--8541.

\bibitem{naowarat23_interspeech}
B.~Naowarat, P.~Harding, P.~D'Alterio, S.~Tong, and B.~{Awwad Shiekh Hasan}, ``{Effective Training of Attention-based Contextual Biasing Adapters with Synthetic Audio for Personalised ASR},'' in \emph{Proc. INTERSPEECH 2023}, 2023, pp. 1264--1268.

\bibitem{shamsian2024keyword}
A.~Shamsian, A.~Navon, N.~Glazer, G.~Hetz, and J.~Keshet, ``Keyword-guided adaptation of automatic speech recognition,'' \emph{arXiv preprint arXiv:2406.02649}, 2024.

\bibitem{liu2020contextualizing}
D.-R. Liu, C.~Liu, F.~Zhang, G.~Synnaeve, Y.~Saraf, and G.~Zweig, ``Contextualizing asr lattice rescoring with hybrid pointer network language model,'' \emph{arXiv preprint arXiv:2005.07394}, 2020.

\bibitem{sun23e_interspeech}
G.~Sun, X.~Zheng, C.~Zhang, and P.~C. Woodland, ``{Can Contextual Biasing Remain Effective with Whisper and GPT-2?}'' in \emph{Proc. INTERSPEECH 2023}, 2023, pp. 1289--1293.

\bibitem{futami2024phoneme}
H.~Futami, E.~Tsunoo, Y.~Kashiwagi, H.~Ogawa, S.~Arora, and S.~Watanabe, ``Phoneme-aware encoding for prefix-tree-based contextual asr,'' in \emph{ICASSP 2024-2024 IEEE International Conference on Acoustics, Speech and Signal Processing (ICASSP)}.\hskip 1em plus 0.5em minus 0.4em\relax IEEE, 2024, pp. 10\,641--10\,645.

\bibitem{le2020g2g}
D.~Le, T.~Koehler, C.~Fuegen, and M.~L. Seltzer, ``G2g: Tts-driven pronunciation learning for graphemic hybrid asr,'' in \emph{ICASSP 2020-2020 IEEE International Conference on Acoustics, Speech and Signal Processing (ICASSP)}.\hskip 1em plus 0.5em minus 0.4em\relax IEEE, 2020, pp. 6869--6873.

\bibitem{le2021deep}
D.~Le, G.~Keren, J.~Chan, J.~Mahadeokar, C.~Fuegen, and M.~L. Seltzer, ``Deep shallow fusion for rnn-t personalization,'' in \emph{2021 IEEE Spoken Language Technology Workshop (SLT)}.\hskip 1em plus 0.5em minus 0.4em\relax IEEE, 2021, pp. 251--257.

\bibitem{gong2024contextual}
X.~Gong, A.~Lv, Z.~Wang, and Y.~Qian, ``Contextual biasing speech recognition in speech-enhanced large language model,'' \emph{Proc. Interspeech. ISCA}, pp. 257--261, 2024.

\bibitem{sun2023contextualbiasingremaineffective}
\BIBentryALTinterwordspacing
G.~Sun, X.~Zheng, C.~Zhang, and P.~C. Woodland, ``Can contextual biasing remain effective with whisper and gpt-2?'' 2023. [Online]. Available: \url{https://arxiv.org/abs/2306.01942}
\BIBentrySTDinterwordspacing

\bibitem{kwok2024continual}
C.~Y. Kwok, J.~Q. Yip, and E.~S. Chng, ``Continual learning optimizations for auto-regressive decoder of multilingual asr systems,'' \emph{arXiv preprint arXiv:2407.03645}, 2024.

\bibitem{kwok2024continual_2}
------, ``Continual learning with embedding layer surgery and task-wise beam search using whisper,'' in \emph{2024 IEEE Spoken Language Technology Workshop (SLT)}.\hskip 1em plus 0.5em minus 0.4em\relax IEEE, 2024, pp. 140--146.

\bibitem{Zhao2019ShallowFusionEC}
\BIBentryALTinterwordspacing
D.~Zhao, T.~N. Sainath, D.~Rybach, P.~Rondon, D.~Bhatia, B.~Li, and R.~Pang, ``Shallow-fusion end-to-end contextual biasing,'' in \emph{Interspeech}, 2019. [Online]. Available: \url{https://api.semanticscholar.org/CorpusID:202716857}
\BIBentrySTDinterwordspacing

\bibitem{Pundak2018DeepCE}
\BIBentryALTinterwordspacing
G.~Pundak, T.~N. Sainath, R.~Prabhavalkar, A.~Kannan, and D.~Zhao, ``Deep context: End-to-end contextual speech recognition,'' \emph{2018 IEEE Spoken Language Technology Workshop (SLT)}, pp. 418--425, 2018. [Online]. Available: \url{https://api.semanticscholar.org/CorpusID:51942169}
\BIBentrySTDinterwordspacing

\bibitem{ortmanns2000look}
S.~Ortmanns and H.~Ney, ``Look-ahead techniques for fast beam search,'' \emph{Computer Speech \& Language}, vol.~14, no.~1, pp. 15--32, 2000.

\bibitem{liu2001new}
F.~Liu, M.~Afify, H.~Jiang, and O.~Siohan, ``A new verification-based fast match approach to large vocabulary speech recognition,'' in \emph{Proc. Eurospeech 2001}, 2001, pp. 851--854.

\bibitem{Radford2022RobustSR}
\BIBentryALTinterwordspacing
A.~Radford, J.~W. Kim, T.~Xu, G.~Brockman, C.~McLeavey, and I.~Sutskever, ``Robust speech recognition via large-scale weak supervision,'' in \emph{International Conference on Machine Learning}, 2022. [Online]. Available: \url{https://api.semanticscholar.org/CorpusID:252923993}
\BIBentrySTDinterwordspacing

\bibitem{kwok2024improved}
C.~Y. Kwok, J.~Q. Yip, and E.~S. Chng, ``Improved alignment for score combination of rnn-t and ctc decoder for online decoding,'' in \emph{International Conference on Text, Speech, and Dialogue}.\hskip 1em plus 0.5em minus 0.4em\relax Springer, 2024, pp. 70--80.

\bibitem{koh2019building}
J.~X. Koh, A.~Mislan, K.~Khoo, B.~Ang, W.~Ang, C.~Ng, and Y.~Tan, ``Building the singapore english national speech corpus,'' \emph{Malay}, vol.~20, no. 25.0, pp. 19--3, 2019.

\bibitem{yuen2023asr}
C.~Y. Kwok, H.~Y. Li, and E.~S. Chng, ``Asr model adaptation for rare words using synthetic data generated by multiple text-to-speech systems,'' in \emph{2023 Asia Pacific Signal and Information Processing Association Annual Summit and Conference (APSIPA ASC)}.\hskip 1em plus 0.5em minus 0.4em\relax IEEE, 2023, pp. 1771--1778.

\bibitem{zen2019libritts}
H.~Zen, V.~Dang, R.~Clark, Y.~Zhang, R.~J. Weiss, Y.~Jia, Z.~Chen, and Y.~Wu, ``Libritts: A corpus derived from librispeech for text-to-speech,'' \emph{arXiv preprint arXiv:1904.02882}, 2019.

\bibitem{kim2021conditional}
J.~Kim, J.~Kong, and J.~Son, ``Conditional variational autoencoder with adversarial learning for end-to-end text-to-speech,'' in \emph{International Conference on Machine Learning}.\hskip 1em plus 0.5em minus 0.4em\relax PMLR, 2021, pp. 5530--5540.

\bibitem{Yamagishi2019CSTRVC}
\BIBentryALTinterwordspacing
J.~Yamagishi, C.~Veaux, and K.~MacDonald, ``Cstr vctk corpus: English multi-speaker corpus for cstr voice cloning toolkit (version 0.92),'' 2019. [Online]. Available: \url{https://api.semanticscholar.org/CorpusID:213060286}
\BIBentrySTDinterwordspacing

\bibitem{le2021contextualized}
D.~Le, M.~Jain, G.~Keren, S.~Kim, Y.~Shi, J.~Mahadeokar, J.~Chan, Y.~Shangguan, C.~Fuegen, O.~Kalinli \emph{et~al.}, ``Contextualized streaming end-to-end speech recognition with trie-based deep biasing and shallow fusion,'' \emph{arXiv preprint arXiv:2104.02194}, 2021.

\bibitem{ravanelli2021speechbraingeneralpurposespeechtoolkit}
\BIBentryALTinterwordspacing
M.~Ravanelli, T.~Parcollet, P.~Plantinga, A.~Rouhe, S.~Cornell, L.~Lugosch, C.~Subakan, N.~Dawalatabad, A.~Heba, J.~Zhong, J.-C. Chou, S.-L. Yeh, S.-W. Fu, C.-F. Liao, E.~Rastorgueva, F.~Grondin, W.~Aris, H.~Na, Y.~Gao, R.~D. Mori, and Y.~Bengio, ``Speechbrain: A general-purpose speech toolkit,'' 2021. [Online]. Available: \url{https://arxiv.org/abs/2106.04624}
\BIBentrySTDinterwordspacing

\bibitem{loshchilov2017decoupled}
I.~Loshchilov and F.~Hutter, ``Decoupled weight decay regularization,'' \emph{arXiv preprint arXiv:1711.05101}, 2017.

\bibitem{kwok2024low}
C.~Y. Kwok, J.~Q. Yip, and E.~S. Chng, ``Low resource language adaptation using two-stage regularization for multilingual asr,'' in \emph{2024 International Conference on Asian Language Processing (IALP)}.\hskip 1em plus 0.5em minus 0.4em\relax IEEE, 2024, pp. 332--337.

\bibitem{kwok2025extending}
C.~Y. Kwok, S.~Li, J.~Q. Yip, C.~Chu, T.~Kawahara, and E.~S. Chng, ``Extending whisper for emotion prediction using word-level pseudo labels,'' in \emph{ICASSP 2025-2025 IEEE International Conference on Acoustics, Speech and Signal Processing (ICASSP)}.\hskip 1em plus 0.5em minus 0.4em\relax IEEE, 2025, pp. 1--5.

\end{thebibliography}

\end{document}